# Evaluation of Protein-protein Interaction Predictors with Noisy Partially Labeled Data Sets

Haohan Wang
Language Technologies Institute
School of Computer Science
Carnegie Mellon University
Pittsburgh, PA, USA, 15213
haohanw@cs.cmu.edu

Madhavi K. Ganapathiraju
Department of Biomedical Informatics
University of Pittsburgh
Pittsburgh, PA 15213
madhavi@pitt.edu

## ABSTRACT

Protein-protein interaction (PPI) prediction is an important problem in machine learning and computational biology. However, there is no data set for training or evaluation purposes, where all the instances are accurately labeled. Instead, what is available are instances of positive class (with possibly noisy labels) and no instances of negative class. The non-availability of negative class data is typically handled with the observation that randomly chosen protein-pairs have a nearly 100% chance of being negative class, as only 1 in 1,500 protein pairs expected is expected to be an interacting pair. In this paper, we focused on the problem that non-availability of accurately labeled testing data sets in the domain of protein-protein interaction (PPI) prediction may lead to biased evaluation results. We first showed that not acknowledging the inherent skew in the interactome (i.e. rare occurrence of positive instances) leads to an over-estimated accuracy of the predictor. Then we show that, with the belief that positive interactions are a rare category, sampling random pairs of proteins excluding known interacting proteins set as the negative testing data set could lead to an under-estimated evaluation result. We formalized those two problems to validate the above claim, and based on the formalization, we proposed a balancing method to cancel out the over-estimation with under-estimation. Finally, our experiments validated the theoretical aspects and showed that this balancing evaluation could evaluate the exact performance without availability of golden standard data sets.

## Categories and Subject Descriptors

D.3.3 [**Artificial Intelligence**]: Learning– *concept learning, knowledge acquisition.*

## General Terms

Algorithms, Performance.

## Keywords

Artificial intelligence, Computational biology, Machine learning, Performance evaluation, protein-protein interaction prediction.

## 1. INTRODUCTION

Protein-protein interaction (PPI) prediction, as a computationally

SPECIFIC TEMPLATE PLACE HOLDER

challenging and bio-medically impactful problem, draws great attentions from both computational scientists as well as biomedical scientists. Computational scientists usually study large-scale prediction of PPIs as a supervised classification problem in machine learning, using numerical features that represent co-localization and co-expression of protein pairs. Solving a supervised classification problem require a statistical model learnt from a dataset with labeled instances. After a suitable model is designed and evaluated on a dataset whose true labels are known, it can be applied to other protein-pairs that promise value in the investment of wet-lab resources. This approach can accelerate biomedical research greatly because such discovery simply by human intuition and scientific reasoning is rarely possible, and typically requires several weeks of labor and investment of material and financial resources to validate a possible PPI; computational prediction of PPI can also be very impactful: as proteins carry out most of the functions in cellular processes, accurate discovery of PPIs can lead to biomedically relevant results. For example, a PPI predicted by Ganapathiraju, et al (unpublished work under review) between the genes OASL and RIG-I has been validated in the lab [29] and led to the knowledge that a naturally occurring protein OASL can boost the viral immunity of an individual, which in turn led to several other results [14] [15] [19].

Verifying computational predictions with experiments on the biomedical side is very expensive; thus, it requires that the computational models be able to predict the interacting pairs with very low Type I error (i.e. having high precision); furthermore, it requires a good mechanism that can evaluate the model with respect to Type I error.

Training and evaluating the algorithms requires labeled data set, and the problem is that there exists no gold standard dataset of 'non-interacting' protein-pairs. This challenge could be conveniently alleviated because the number of interacting pairs is estimated to be only one in about 300 to 1,500 protein pairs, thereby allowing randomly chosen pairs to surrogate as negative data in PPI prediction [8] [16] [22], but a slight biased evaluation is introduced. Another notable fact is that evaluation testing data set should follow the natural skewness, that a dominating majority of protein pairs are negative (non-interacting) pairs. Failure to acknowledge this skewness in test data set will also lead to a biased evaluation result, as we will discuss in detail later in this paper. Given that the development of interactome-scale PPI prediction algorithms is on the rise, it is important to calibrate how these biased evaluation result will affect the understanding of a model by computational scientists.

To study the performance of these biased evaluations for PPI prediction models at interactome scale, we first formalize the expected Type I error under the influence of the skewness of the data set. We showed that both over estimation and under estimation could occur. Then we propose a method to cancel out these over estimation and under estimation, thus to get a more accurate evaluation result of a model.

To validate our formalism, we carried out experiments on yeast interactome data set. Yeast has 1 in 230 pairs to be known interactions which is by far the most known PPIs in relation to the estimated size of its interactome [1]. Because interactome dataset of yeast is more extensively studied than the one of human, we simulate the characteristics of the interactome prediction with the yeast dataset to comprehensively evaluate models and see if the evaluations are biased.

## 2. RELATED WORK

The problem that non-availability of golden negative interacting pairs may lead to biased evaluation of PPI prediction model has drawn attention from many different researchers from different angles.

In [20] Park et al clarify the difference between two types of subset sampling for negative PPIs: 1) sampling for cross-validated testing set where unbiased subsets are desired so that predictive performance estimated with testing set can be believed to generalize to the population level; 2) sampling for training set, where whether the sampled bias is not a problem as long as a good predictive model can be trained out of it. They also showed that balanced sampling proposed by [28] is a suitable method for sampling training set, but not testing set. Similarly, [11] has showed this problem empirically and focused on the problem raised by cross-validating set. They suggested that the assessment of performance should be accomplished through many other traditional statistics and machine learning scenario, like 1) training and testing with different datasets of different interpretations of negative interacting pairs, 2) generating other datasets with a combination of the original datasets and 3) evaluating through a wider range of statistics.

Instead of looking into this problem statistically, other researchers tried to solve this problem by incorporating some domain knowledge into the creation of data set [18] [23] [24] has constructed a data set with the protein pairs that have a high probability to be not interacting pairs. In [18], they make the assumption that proteins within different subcellular localizations will not interact; so non-interacting pairs could be sampled as long as each protein of that pair is from a different localization. In [23], they hold the belief that certain proteins have been studied more extensively than other proteins. Therefore, instead of random sampling, they selected the negative data set composed of proteins (but not protein pairs) from the positive interacting data set. However, the problem of this sophisticated sampling approach is that such a sampling method will statistically simplify and mis-direct the machine learning process because the data their model faces has been separated with prior knowledge. It is very likely that eventually, the model learns the prior knowledge used to sample the data, instead of the underlining statistical pattern researchers are interested in. Because the test data is sampled in the same way, it is not trivial to detect such an overfit with testing data.

Another interesting approach is to take a detour of the problem of lack of gold-standard negative data. Instead of trying to sample out negative data or to use the negative data selected with prior knowledge, some researchers decide to work on the computational models. Many models have been proposed to work with only positive labeled data set [9] [17] and some of the models have been proposed specifically to bioinformatics domain, like [5] [25] [27]. These computational models are developed based on the idea of Support Vector Machines, to learn certain information for a specific application with only positive labeled data. However, the problem of how to evaluate these computational models with non-availability of negative data set still remains the same.

In this paper, as an extension to the previous work focusing on the problem of non-availability of negative interacting pairs, our work is trying to solve the evaluation part of it with a different approach. First of all, instead of showing this problem empirically, we formally analyze the problem and showed the bias between exact performance and performance evaluated by current sampling method. Then, based on the formalism, we take a step further to design an exact evaluation method to allow us evaluate the performance of the predicting model with only randomly sampled pairs, which we will show in the next section.

## 3. EXACT EVALUATION INFERED FROM BIASED EVALUATION

### 3.1 Over-estimation without Skewness of Data

As a start, we assume the species we are working on is estimated to have 1 in $P$ pairs to be positive class (interacting pairs). To make it simple, we use $p$ to denote the fraction of all the estimated interacting pairs over all the possible pairs, thus we have $p = 1/P$.

Let $\alpha$ be the accuracy of the model we build. First, we assume the model functions equivalently for distribution of positive and negative accuracy classes, i.e. the accuracies of the model on either positive data or negative data are $\alpha$. We will relax this assumption later.

Thus, for a test data set with N protein pairs, with a manually selected $\tilde{p}$, our model will predict $\alpha \tilde{p} N + \varepsilon_1$ true positive intracting pairs and $(1 - \alpha)(1 - \tilde{p})N + \varepsilon_2$ false interacting pairs. Here $\varepsilon$ stands for predicted protein pairs affected by the unpredictable natural noises with data.

With equivalence of working with Type I error, we focus on precision. Now, we can achieve a precision score as the following:

$$precision = \frac{\alpha \tilde{p} N + \varepsilon_1}{\alpha \tilde{p} N + (1-\alpha)(1-\tilde{p})N + \varepsilon_1 + \varepsilon_2} \qquad (1)$$

We want to study how $p$ could influence the precision score, thus, we have the following:

$$precision = \frac{\alpha + \frac{\xi_1}{\tilde{p}}}{2\alpha - 1 + \frac{1-\alpha}{\tilde{p}} + \frac{\xi_1 + \xi_2}{\tilde{p}}} \qquad (2)$$

in which $\xi = \varepsilon/N$.

If we discard the noise terms here, we can simply see that $precision \propto \tilde{p}/(C + \tilde{p})$ ($C$ is a constant). Thus, testing Type I error of the model with a non-realistic data set will non-realistic evaluation score.

Further, we relax the assumption constrains the model work similarly over positive interacting pairs and negative ones. We use

$\alpha_1$ and $\alpha_2$ to denote the accuracy of the model over two classes, and we have $\alpha = \alpha_1\tilde{p} + \alpha_2(1-\tilde{p})$. The precision derived from here is not very different from the equations above, we have:

$$precision = \frac{\alpha_1 + \frac{\xi_1}{\tilde{p}}}{\alpha_1 + \alpha_2 - 1 + \frac{1-\alpha_2}{\tilde{p}} + \frac{\xi_1 + \xi_2}{\tilde{p}}} \qquad (3)$$

Here we can draw the same conclusion that $precision \propto \tilde{p}/(C' + \tilde{p})$ and the proportion of positive interacting pairs will affect the evaluation score in the same way.

Usually, with traditional machine learning customs, computational scientists prefer to set the test data with almost balanced distribution of positive and negative protein pairs, where $\tilde{p}$ in the above equations is about 50%. Thus $\tilde{p} \gg p$, and this practice exaggerates the evaluation score dramatically over the $p$, which is only less than 1%. This gives one of the reasons that computational scientists keep claiming that they have solved the PPI prediction problem [3] [4] [6] [10] [13] [23] [26], but biomedical scientists are unwilling to invest their resources to validate the predicted interactions in the lab despite the high reward if the predicted PPI is indeed found to be correct.

## 3.2 Under-estimation with Neglecting Positive Interactions in Random Pairs

Now we focus on the second biased estimation problem to see the effects of treating all the randomly sampled pairs as negative class (non-interacting pairs).

Again, we assume the species we are working on is estimated to have 1 in $P$ pairs to be positive class, and we have $p = 1/P$. Further, we have a total number of $T$ possible protein pairs, and the interacting pairs are estimated to be the number of $pT$, among which, $K$ interacting pairs are already discovered. Thus, there are still $pT - K$ interacting pairs to be discovered. We use $q$ to denote the fraction of potential interacting pairs to be discovered in the set of $T - K$, and we have $q = (pT - K)/(T - K)$.

Again, we start with the assumption that the model functions similarly over positive and negative classes. Again, we assume the accuracy is $\alpha$.

With a test data set of $N$ protein pairs, with a manually designed $\tilde{p}$, and an inherent $q$, our model can predict $\alpha\tilde{p}N + \alpha(1 - \tilde{p})qN + \varepsilon_1$ true positive pairs and $(1 - \alpha)(1 - \tilde{p})N + \varepsilon_2$ false positive pairs. Thus, we have the precision derived as following:

$$precision = \frac{\alpha\tilde{p}N + \alpha(1-\tilde{p})qN + \varepsilon_1}{\alpha\tilde{p}N + (1-\alpha)(1-\tilde{p})N + \varepsilon_1 + \varepsilon_2} \qquad (4)$$

From here, we can easily see that precision should be increased compared to Equation (1). Thus, we can draw the conclusion that under this assumption, treating all the random protein pair as non-interacting pairs will result in an under-estimation.

This assumption can be relaxed as trivially as what we have in the previous case. With $\alpha_1$ and $\alpha_2$ denoting the accuracy on positive class and negative class respectively, our model now predict $(1 - \alpha_2)(1 - \tilde{p})N - \alpha_1(1 - \tilde{p})qN + \varepsilon_2$ false positive pairs. Here, we notice that the term appended is a function of $\alpha_1$ instead of $\alpha_2$, because the model captures the underlining distribution instead of its appearing label. (The model does not even know the label). Thus, in the denominator, two appended terms with $q$

cancel out each other. As a result, precision is almost the same as in previous case:

$$precision = \frac{\alpha_1\tilde{p}N + \alpha_1(1-\tilde{p})qN + \varepsilon_1}{\alpha_1\tilde{p}N + (1-\alpha_2)(1-\tilde{p})N + \varepsilon_1 + \varepsilon_2} \qquad (5)$$

Therefore, the same conclusion is drawn that neglecting the interacting pairs in randomly sampled data will result in under-estimation.

However, we can trivially know that $q < p$, where $p$ is already a very small portion that is around 1/300. Even $q$ is only slightly smaller than $p$ (because K << T), discarding the influence of $q$ will not impact a lot on the evaluation results. That is the assumption followed currently in PPI prediction research.

## 3.3 Balancing out Over-estimation and Under-estimation

From the above two sections, we can see that for a PPI prediction model, we desire to evaluate the exact precision which could be achieved by:

$$precision = \frac{\alpha pN + \varepsilon_1}{\alpha pN + (1-\alpha)(1-p)N + \varepsilon_1 + \varepsilon_2} \qquad (6)$$

However, we usually end up with a biased evaluation given in Equation (5). In this section we aim to define a $\tilde{p}$ where by manually selecting $\tilde{p}$ higher than $p$, we could be balancing out the under-estimation of $q$ with this over estimation of $p$. The balancing out procedure is formalized as minimizing the following term with respect to $\Delta p$

$$(\frac{\alpha + \alpha q\frac{1-p-\Delta p}{p+\Delta p} + \frac{\xi_1}{p+\Delta p}}{2\alpha - 1 + \frac{1-\alpha}{p+\Delta p} + \frac{\xi_1+\xi_2}{p+\Delta p}} - \frac{\alpha + \frac{\xi_1}{p}}{2\alpha - 1 + \frac{1-\alpha}{p} + \frac{\xi_1+\xi_2}{p}})^2 \qquad (7)$$

with:

$$\tilde{p} = p + \Delta p$$

We term $\xi$ here. The problem could be solved by simple calculus by taking derivative of (7) with respect to $\Delta p$. By setting the derivative term to zero, we can get the function as following:

$$\Delta p = h(q) = \frac{m(p-1)q}{mq + \alpha - 1} \qquad (8)$$

in which:

$$m = (2\alpha - 1)p + 1 - \alpha$$

Thus, $\Delta p$ could be determined by the prior knowledge of the species. Once we have an appropriately selected $\Delta p$, we could perform exact evaluation over the PPI prediction model.

Now, with Equation (8), we have shown that by selecting the $\Delta p$ wisely, we can get the exact evaluation result even when the gold standard data set is not available.

It seems that there is still a problem. In Equation (8) $\Delta p$ depends on $\alpha$, which is a variable we will never know exactly unless we can evaluate our model first. However, this problem can be easily solved because of the relationship between $\Delta p$ and $\alpha$. Figure 1 shows $\Delta p$ as a function of $\alpha$ for $0 < \alpha < 1$, with different values of $p$. Here, since $p$ is the fraction of all the estimated interacting pairs over all the possible pairs, we only consider the $p$ ranging from 1/ 100 to 1/ 1000 and we set $q$ to be $p/100$. From Figure 1, we can see that the value of $\Delta p$ doesn't change much unless the model has a very high $\alpha$. However, we will never expect our model to have a very high $\alpha$ because of the fact that there will be interacting pairs sampled as non-interacting pairs in both training set and testing set, the model will hardly predict PPI with a 95% or higher accuracy.

With this reasoning, we can safely choose an $\alpha$ and use this $\alpha$ to derive the representation of $\Delta p$ as a function of $p$ and $q$. Here, to simply the representation, we assign $\alpha$ as 1/2, thus, we have:

$$\Delta p = \frac{(1-p)q}{1-q}$$

where $p$ is the believed fraction of interacting protein pairs out of all protein pairs and $q$ is the fraction of potential interacting pairs to be discovered, as formally defined in Section 3.2. We believe that by setting $\tilde{p} = p + \Delta p$, we will be able to get an almost exact evaluation of model performance even without the full knowledge of data.

To derive the representation of $\Delta p$, we introduce an approximation of treating all possible values of $\alpha$ equally. However, this approximation can barely affect our approximation because the variance of $\Delta p$ is upper bounded with $1e^{-6}$ for different choices of $\alpha$ when $0 < \alpha < 0.95$.

In this section, we have formalized the problem of biased evaluation with lack of knowledge of gold standard non-interacting pairs and we have proposed a balanced method to accurately evaluate the PPI predicting model. In the next two sections, we will first validate our formalism with experiments on the real yeast data set then we will prove the feasibility of our exact evaluation method with artificial data set.

# 4. EXPERIMENTAL RESULTS ON YEAST INTERACTOME DATA SET

## 4.1 Experiment Set Up

### 4.1.1 Data
The yeast interactome was downloaded from the Saccharomyces Genome Database (SGD) [7]. It contains 6,325 genes and 125,898 biophysical interactions. Out of these, 82,593 were unique non-self-physical interactions. Sequence based features were used to predict PPIs; therefore only the genes whose protein sequences are available, are used in this work. This includes 5,866 genes and 82,593 unique interactions. To create a partial interactome for studies described here, 20 percent of interacting proteins are sampled randomly from the data set and there are 5,008 interacting pairs for these proteins.

### 4.1.2 Simulation
Figure 2 shows a Venn diagram of the data sets of interest in our experiment: total number of protein-pairs ($\mu$), estimated number of interacting pairs ($A$), number of known interacting pairs ($B$), and the typical overlap of training and evaluation datasets. Out of all the species, set A is the most likely to be completely known in yeast because yeast is the most extensively studied specie. In other words, set B is the most likely to be the same as set A in yeast data set, whereas, in most other organisms a very small subset, which is represented by Set B, is known. To simulate the case of other organisms, an equivalent of set B for yeast is created

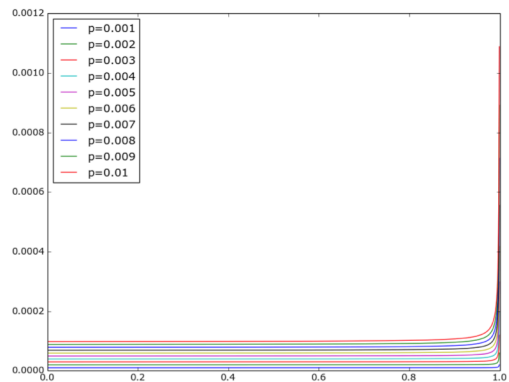

**Figure 1. $\Delta p$ as a function of $\alpha$ for different possible values of $p$**

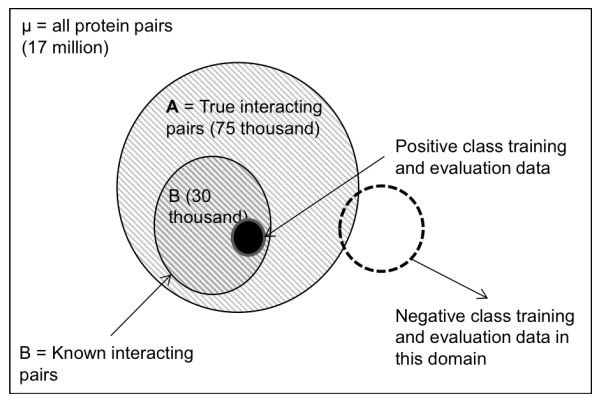

**Figure 2. Venn diagram showing the overlap of negative-labeled data in training and tests with interacting and non-interacting protein pairs. $\mu$ is the set of all protein-pairs of an organism, A is the set of all interacting pairs, B is the set of known interacting pairs. $\mu \setminus A$ is the real set of non-interacting pairs, but typically randomly data from $\mu \setminus A$ is treated as non-interacting pairs for training and testing of PPI prediction algorithms.**

by sampling 20 percent of proteins from set A, out of which there are 5,008 interacting pairs.

### 4.1.3 Features
The features we use mainly fall into two categories. For the first category, we have co-expression, Gene Ontology (GO) [2] similarity and MIPS [18] similarity which are downloaded from Gerstein laboratory[1]. In the second category, we use sequence based features, like frequency of optimal codons, hydropathicity of protein, and frequency of aromatic amino acids, downloaded from SGD yeast database[2].

### 4.1.4 Algorithm
Random Forest models are widely used for PPI prediction and it has been proved to be the machine learning method with best performance [21]. RF implementation from the Weka library was used in this work [12]. Number of random features considered at each node was set to 3, which corresponds to the integer value of

---

$log_2$ (total number of features)+1 as it is the recommended number; number of trees was set to be 30.

## 4.2 Experimental Results

### 4.2.1 Evaluation with non-skewed test data

First, in order to validate the claim that failing to admit the fact that positive interacting pairs are rare cases out of all interacting pairs will lead to over-estimation of model performance, we start with all correctly labeled data. We created a training set of size 100,000 with the positive to negative ratio being 1:4, where positives are selected from all known interactions (set A) and negatives from all other possible pairs ($\mu \setminus A$).

Four test sets were created with $\tilde{p} = 0.4\%$, 1%, 5% and 10%. It is observed that the performance is exaggerated when a larger proportion of positive instances is used, as opposed to using only 0.4% positives, as shown in Figure 3. This validates our belief that the performances of the models are over-estimated when the positive pairs in testing data are more than its in the natural data set, thus, results could be significantly exaggerated when evenly split test data are applied.

### 4.2.2 Evaluation with typical test data

Now, we introduce the problem caused by the existence of $q$, as introduced in Section 3.2. In order to simulate the scenario with the existence of non-discovered interacting pairs, positive instances were taken from set B and negative instances from $\mu \setminus B$ to create train and test data set.

For the sake of clear comparison, we keep the size and positive negative ratio of train and test data the same as the train and test data in the Section 4.2.1. The result is shown in Figure 4.

Similar to Figure 3, Figure 4 also shows a trend of decreasing performance with increasing skew in the number of positive and negative instances. There is not much noticeable difference between Figure 3 and 4 for data with 0.4% and 1% positive instances, but we could see some difference for data with 5% and 10% positive instances. We could tell that the performance evaluation of predicting model is affected by the existence of $q$, but not dramatically.

These observations lead to the next question: how much influence do incorrect labels in training data have on the accuracy of the model, as opposed to incorrect labels in test data. The labels in the test sets were corrected to their true labels and the performance of the algorithm was re-evaluated. That is, the test sets now have correct labels whereas training set still contained noisy labels for negative set. The results were shown in Figure 5 in comparison to the corresponding lines from Figure 3 and 4. It is evident that when the test data labels are corrected, the performance is comparable to the case when both training and test data had correct labels, as in Figure. 3. This shows that random forest was considerably robust against noise in the training data, but the non-availability of real negative labels in the test data grossly under-estimates the true capability of the algorithm in real application where the data has a skewed distribution.

To measure the difference that may not be recognized in the figure, we calculated the area under precision recall (PR) curve (see Table 1). In the table, different test data results were seen in different area under PR curve. It shows that when the label is corrected, the score is higher than what can be actually achieved through the experiment setting. This suggests that our current strategy of PPI evaluation is an under-estimate of the classification model.

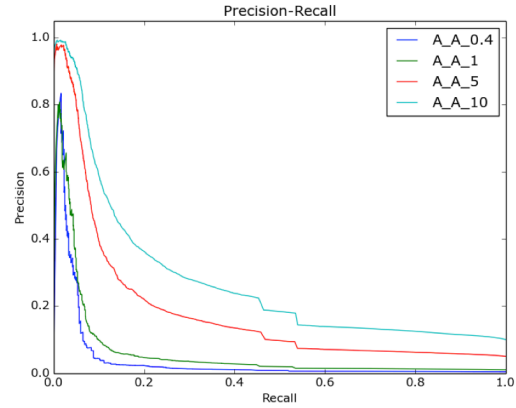

**Figure 3. A prediction model is trained with training and testing data that are correctly labeled. Results are shown at 4 difference mixing ratios of positive class in overall data: 0.4%, 1%, 5% and 10%.**

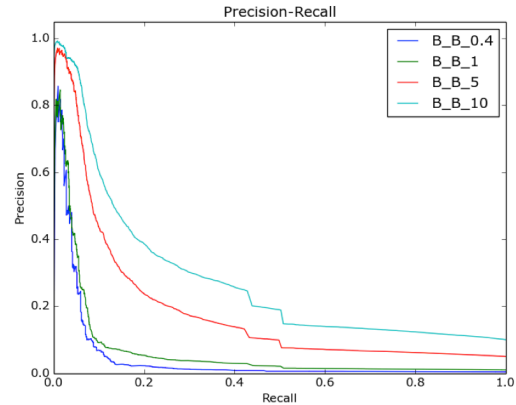

**Figure 4. A prediction model is trained with training and testing data that whose negative instances have partial overlap with true positives. Results are shown at 4 difference mixing ratios of positive class in overall data: 0.4%, 1%, 5% and 10%.**

**Table 1. Area under precision recall curve for model tested by different data sets.**

| Test Data set | 0.4% of positive | 10% of positive |
|---|---|---|
| Sample from B set | 0.044 | 0.279 |
| Sample from A set | 0.046 | 0.280 |
| Corrected Label | 0.050 | 0.284 |

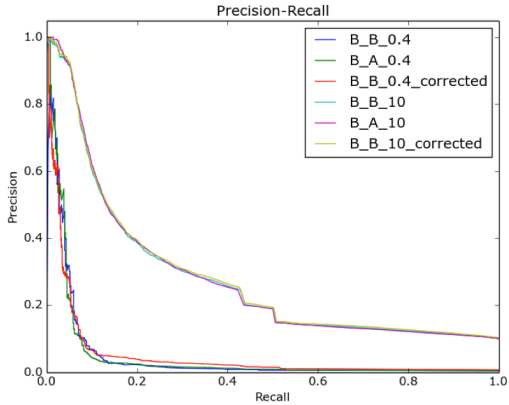

**Figure 5. Precision-recall curves are recomputed from results corresponding to previous figure by assigning correct labels to negative instances of test data alone, and re-computing the observed precision and recall. Precision improves at any given recall because some of the predicted interactions that previously had negative labels are now found to be positive labels.**

In this section we have set up the experiment in real yeast data set to validate our claim about the biased evaluation resulting in partially known data set. Along our way of showing that, we also showed the robustness of Random Forest for the noises of training data. However, since we do not have authentic information of the property of the nature data set, like $p$ and $q$, we can only validate our proposed evaluation method with artificial data set. We will show the detailed experiment in the next section.

# 5. EXPERIMENTAL RESULTS ON ARTIFICIAL DATA SETS

## 5.1 Experiment Set Up

Our artificial data for positive and negative class are generated from two different 100-dimension multivariate Gaussian distribution with slightly different centroids and same variance. We generate the data in the ratio that for 100 data instances, there will be only 1 instance is positive. Therefore, the generated data set basically captures the two critical features of our protein interacting data set. It is very noise, thus difficult to distinguish between positive data and negative data, and positive data are rare.

Following the tradition in last section, we set current positive data set as Set A, and then sampled 20% positive data out of Set A as the Set B.

To be consistent, we selected Random Forest as the classifier for this section and we set the parameters the same as last section.

## 5.2 Experimental Results

With the exact same experiment set up in the previous section, we have performed the experiment on the artificial data set. Figure 6 shows our experimental result, with the precision-recall curve calculated with when all the positive and negative data are available ("ideal" case), when only partially data are available ("real" case) and when our method is applied to the real case ("proposed" case) respectively.

Although our data is generated following the positive negative ratio of 1:99, we still perform our experiment for four different portions of positive data, to be consistent with the experiment in the previous section. Same as in previous section, these four sets of experiment are performed for the percentages of positive data are 0.4%, 1%, 5% and 10% respectively. These four plots are showed from left to right in Figure 6.

One important aspect to notice for the results in this section: our generated data can only simulate the real PPI data set qualitatively because the features describing the data are only generated from standard distributions, like Gaussian distribution, Poisson Distribution and Beta distribution. We are unable to generate the features according to the relationship between real world features and the protein pairs because such a relationship is one the fundamental problems both computational scientists and

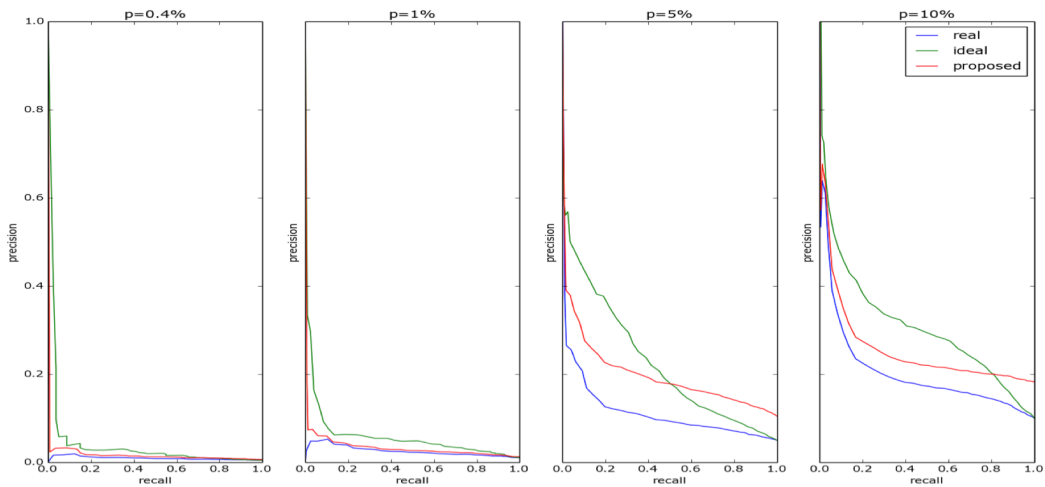

**Figure 6. Precision-recall curve for three different cases ("real" in blue, "ideal" in green, "proposed method" in red) across four different experiment settings (from the left to right, the percentage of positive interacting pairs are 0.4%, 1%, 5% and 10% respectively)**

biologists are trying to solve.

From Figure 6, we can see that there is still a gap between our proposed evaluation method and the ideal case. However, the gap between our method and the ideal case is smaller than the gap between real world cases and the ideal case. This difference is consistent across all the 4 different settings for the experiment. We can get the conclusion that, although our proposed still cannot get the exact evaluation result as our equations showed, we can have a better evaluation result than what we used to have now.

For a clearer contrast of these three cases across the four different experiment settings, we have calculated the area under the precision recall curve and showed that in Table 2.

**Table 2. Precision recall curve for three cases and four experiment settings.**

|          | 0.4%  | 1%    | 5%    | 10%   |
|----------|-------|-------|-------|-------|
| Real     | 0.009 | 0.025 | 0.113 | 0.202 |
| Proposed | 0.019 | 0.035 | 0.167 | 0.224 |
| Ideal    | 0.041 | 0.05  | 0.223 | 0.308 |

As we can see from the table, our proposed method has a better approximation towards the real case when $p$ is set to 0.4% and 1% than when $p$ is set to 5% and 10%. We believe the reason is that our conclusion in Section 3 is drawn when $p$ is set to 1%. Different conclusions could be drawn is $p$ is significantly larger than 1% because, as one of our critical step, $\Delta p$ will not be affected by $\alpha$ is only valid when $p$ is small (thus $q$ is even smaller) compared to 1.

One of the reasons that our proposed method cannot evaluate the model exactly may be that we have lost some information when we only use one real number to approximate $\alpha$. Another reason could be because that our generated data may be even noisy than real world PPI data set, thus the gap is due to the noises of the data.

In this section, we have validated our proposed exact evaluation method with some generated data set. We have shown that, although our proposed method cannot evaluate the model exactly, it still shows some improvement over the traditional evaluation method.

# 6. PREDICTION AT WHOLE-INTERACTOME SCALE

Besides the exact evaluation methods we have proposed and validated, here, we would like to introduce another way of evaluating the PPI predicting model. This evaluation method has been empirically shown to be working well and it is unlikely to suffer from the problem of lack of fully-annotated data set. The key idea is similar to [10] [23] [24], because we all focus on validating every pair of non-interacting protein pairs to be confidently non-interacting. However, what we proposed here has two major advantages over their work: 1) As we mentioned in Section 2, we only consider the prior-knowledge-based data in the testing phase, thus we can guarantee that the model we trained learns the general statistical pattern of the data set. 2) Instead of sampling a subset that may or may not be meaningful in real world, we propose to use hub proteins directly.

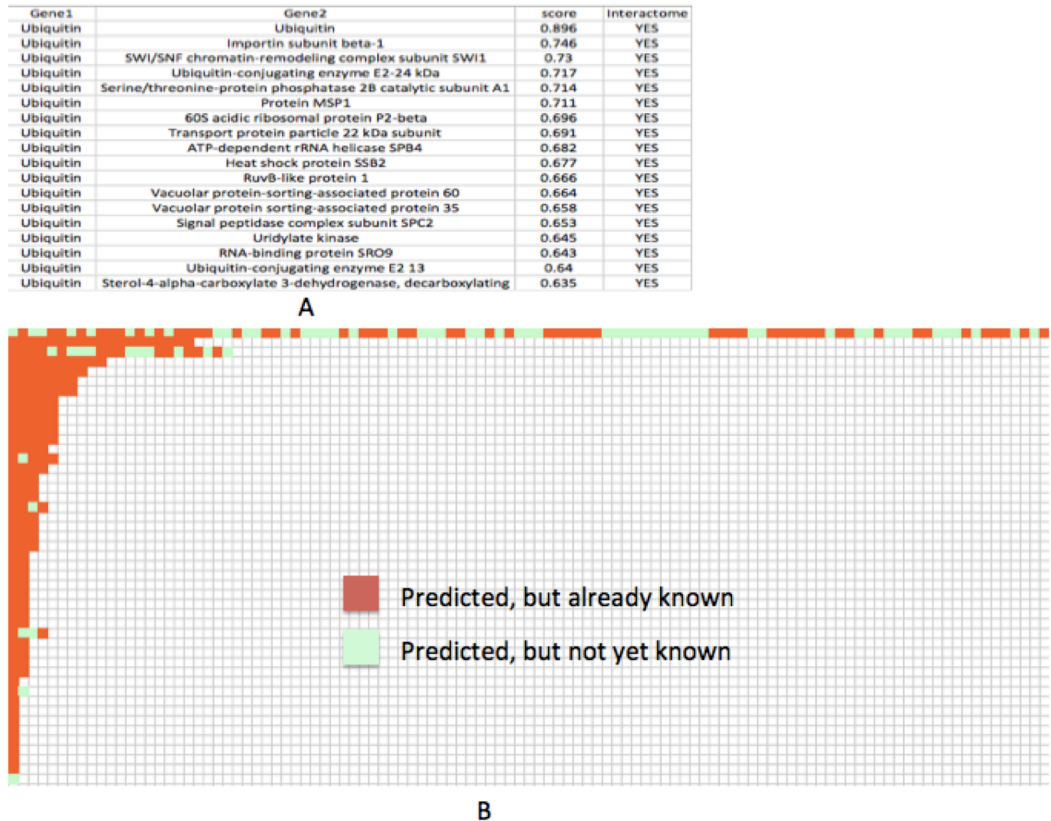

**Figure 7. (A) Ranked list of predicted interactions are shown for a human hub protein, Ubiquintin. Predicted interactors (column 3) are ordered in descending order of predicted score (column 2). If the predicted interaction is already known to be interacting with Ubiquintin, it is indicated in the last column 'isKnown' with a yellow color and marked "yes". For an accurate model, typically all the top predictions have a "Yes" in the isKnown column. (B) All the prediction result is shown, where each row stands for each protein of interest and the columns stand for the set of interactions predicted. Known interactions are shown in yellow color and unlabeled pairs are shown in blue color. Note that, the known interactions are the positive instances from set B.**

We believe that hub proteins (i.e. proteins with more than 50 known interactions) serve as nearly fully-annotated test data. This is because, hubs tend to be very well studied, and given that many of their interactions are already documented, we believe that most, if not all, of their interactions are already known. The ranked lists of hub proteins may be parsed downwards from top, to check up to what threshold the predictions yield predominantly known interactions. If the model is accurate, there should be only a few, if any, unlabeled pairs towards the top in these ranked lists. The fourth column in Figure 7A which indicates whether the prediction is a true interaction, is shown for all the hub genes in Figure 8B; with a chosen threshold on score, all predictions above the threshold are considered to be 'predicted interactions'. The cells are colored orange for predicted interactions that are already known and green, if not. For an accurate prediction model, most of the predicted genes ranked higher would be true interactions. The few high confidence unlabeled pairs (predicted interactions that are not known interactions, such as first prediction of RSP13 (in the first row) or first two for HHF2 (in Figure 7B) could indeed be true interactions that were hitherto not studied by experiments. A threshold may be chosen for a desired performance, and then applied to other proteins with the belief that the predictions above the threshold are accurate. This method of evaluating the PPI predicting model, although has not been fully proved with the feasibility, was shown to be an effective method for comparing interactome-scale predictions and for consideration by biologists (unpublished results currently under review).

## 7. CONCLUSIONS

In this work, we visited the problems caused by the non-availability of golden standard negative data set in bioinformatics community again. Here, we mainly focused on the problem for the application of protein protein interaction prediction. We mainly focused that such a non-availability will impact the evaluation of PPI predicting model and it may cause either over-estimation or under estimation for different scenarios.

We first formalized the problem we are trying to solve and showed that how the over-estimation and under-estimation could happen. Then, with the formalization, we took a step further to intentionally cause over-estimation and under-estimation. With a dedicatedly selected amount biased estimation, we are aimed to balance off the biases from both directions. With some other assumptions and approximations, we derived an exact evaluation formula.

In the first experiment, we made use of the yeast interactome data to study the effects of labeling random data as negative. We see that results in the PPI domain are closer to the real world when reported on datasets with higher imbalance. Reporting results on datasets that are not as skewed as in the human interactome seem to exaggerate the performance of algorithms. Using random instances as negative instances underestimates the true capability of the algorithm, although such noise in training data does not affect the performance much.

We believe that our evaluation of the impact of nonavaliability of golden standard data will direct the approaches of evaluating PPI predictions models. Experiments results reported with balanced test data set are not so meaningful. Considering random protein pairs as negative, though not the ideal situation, does not affect the evaluation result very much.

Then, with artificial generated data set, we validate our proposed method. We have shown that our proposed method can evaluate the predicting model more accurately towards the result evaluated by gold standard data compared to traditional evaluation setting. However our method still cannot evaluate the model exactly as evaluation done with golden standard data set, maybe due to the approximation we made or the noises of the data set.

Besides our method, we also proposed and demonstrated that hub proteins can be used to assess precision and recall and to select a threshold for a desired tradeoff. By selecting an appropriate threshold with inspectation to hub proteins, the trade off of precision and recall could be balanced. Thus, a model could be evaluated by appropriate observations on the predicting results of hub proteins. Although this method is lack of scientific provement, it has been shown as an effective method empirically.

## 8. ACKNOWLEDGMENTS

The authors thank Lavanya Viswanathan for data collection and preprocessing. This work has been funded by the Biobehavioral Research Awards for Innovative New Scientists (BRAINS) grant R01MH094564 awarded to MKG by the National Institute of Mental Health of National Institutes of Health (NIMH/NIH) of USA.